\documentclass[letterpaper]{article}
\usepackage{aaai}
\usepackage{times}
\usepackage{helvet}
\usepackage{courier}

\usepackage{graphicx}
\usepackage{color}
\usepackage{algorithm}
\usepackage[noend]{algpseudocode}
\usepackage{xspace}
\usepackage{amsmath}

\DeclareMathOperator*{\minBelow}{min}

\DeclareMathOperator*{\maxBelow}{max}
\usepackage{amssymb}

\newcommand{\MethodName}{MDGAN\xspace}

\graphicspath{ {images/} }
\begin{document}
%
\title{MDGAN: Boosting Anomaly Detection Using \\Multi-Discriminator Generative Adversarial Networks}
\author{Yotam Intrator, Gilad Katz, Asaf Shabtai\\
Department of Software and Information Systems Engineering\\
Ben-Gurion University of the Negev\\
Be'er Sheva, Israel\\
}
\maketitle

\section{ABSTRACT}

Anomaly detection is often considered a challenging field of machine learning due to the difficulty of obtaining anomalous samples for training and the need to obtain a sufficient amount of training data.
In recent years, autoencoders have been shown to be effective anomaly detectors that train only on "normal" data.
Generative adversarial networks (GANs) have been used to generate additional training samples for classifiers, thus making them more accurate and robust.
However, in anomaly detection GANs are only used to reconstruct existing samples rather than to generate additional ones. 
This stems both from the small amount and lack of diversity of anomalous data in most domains.
In this study we propose \MethodName, a novel GAN architecture for improving anomaly detection through the generation of additional samples. 
Our approach uses two discriminators: a dense network for determining whether the generated samples are of sufficient quality (i.e., valid) and an autoencoder that serves as an anomaly detector. 
\MethodName enables us to reconcile two conflicting goals: 1) generate high-quality samples that can fool the first discriminator, and 2) generate samples that can eventually be effectively reconstructed by the second discriminator, thus improving its performance. 
Empirical evaluation on a diverse set of datasets demonstrates the merits of our approach.

\section{INTRODUCTION}
\label{sec:introduction}

In machine learning, anomaly detection aims to identify abnormal patterns, particularly those that arise from new classes of behaviors \cite{chandola2009anomaly}. 
Although it has been extensively researched, anomaly detection is often considered a challenging field of machine learning due to the difficulty of obtaining anomalous samples for training, a problem that often results in a high false positive rate.

In recent years, deep neural nets (DNNs) have been used for training anomaly detection models \cite{kwon2017survey}.
Despite their ability to learn complex patterns and potentially generate accurate anomaly detectors, deep learning models often require large amounts of training data \cite{krizhevsky2012imagenet}. Moreover, samples of anomalous data must be provided to the algorithm in order to enable it to detect similar samples. Obtaining such samples can be difficult, and defining all possible types of anomalies is often close to impossible.

Generative Adversarial Networks (GANs) \cite{goodfellow2014generative} have been proposed in recent years as a solution to the two above-mentioned problems. GANs have been used both to generate additional labeled samples \cite{odena2016conditional} and to make classifiers more robust to adversarial attacks \cite{lee2017generative}. However, to the best of our knowledge, no GAN-based solution has been proposed for generating additional samples in the domain of one-class anomaly detection. The likely cause for the lack of research in this area is the difficulty for the GAN to generate ``anomalous'' samples when only "normal" ones are available for training (i.e., it is impossible to generate samples from all participating classes).

Another obstacle to using GANs in anomaly detection is the requirement in some domains that the generated samples be \textit{valid}. This requirement exists in fields such as network-based intrusion detection, where the generated samples need to be realistic in order not to "throw off" the detection algorithm. In cases where valid samples are not a prerequisite (e.g., image classification), existing solutions can be found in the literature \cite{frid2018synthetic,lemley2017smart,shrivastava2017learning}. 

In this study we aim to improve the performance of anomaly detection algorithms, specifically autoencoders \cite{sakurada2014anomaly}, by using GANs for generating new artificial examples of "normal" cases. 
In order to achieve this goal, we present Multi-Discriminator GAN (\MethodName), a novel GAN architecture that uses two discriminators, each with a different role and cost-function. 
The first discriminator attempts to discern the generated samples from the original ones, thus ensuring that that the generated samples appear as if they are sampled from the same distribution as the real data (i.e., valid). 
The second discriminator is an autoencoder that serves as an anomaly detector by measuring reconstruction error. 

\MethodName's use of two discriminators enables the generator component to achieve two seemingly conflicting goals: 1) generate high-quality samples that can fool the first discriminator, and 2) generate samples that can be reconstructed effectively by the second discriminator (the autoencoder). 
The proposed setting prevents \MethodName from generating simplistic samples that would be easily reconstructed, thus forcing the autoencoder to continuously improve its performance.

To evaluate the merit of our proposed approach we conducted an empirical analysis on ten datasets of varying domains and characteristics (e.g., number of samples, number of features, etc.). The results of our analysis show that \MethodName outperforms a widely-used benchmark in the large majority of tested datasets.

The contributions of this study are twofold: (1) we propose a novel GAN architecture that enable the generation of more finely-tuned training samples for one-class anomaly detection, and; (2) we present an in-depth analysis of the performance of our proposed approach and its components.



\section{RELATED WORK}

\subsection{Anomaly Detection}

Anomaly detection algorithms focus on finding patterns that do not conform to expected behavior. 
Anomaly detection has been applied in various areas, including the fields of fraud detection \cite{van2015apate}, cyber security \cite{kuypers2016empirical}, medicine \cite{james2014medical}, and even real-time crime detection \cite{ravanbakhsh2017training}. 

In this study we focus on spectral anomaly detection methods \cite{egilmez2014spectral}. Approaches of this type focus on generating a lower-dimensionality (i.e. compressed) representation of the data, and then using it to reconstruct the original data. 
The underlying logic is that high reconstruction error is indicative of an anomaly, since the characteristics of the original data are not "as expected." 
This approach includes algorithms such  as principal component analysis (PCA) \cite{shyu2003novel} and autoencoders \cite{sakurada2014anomaly}. 
In this study we focus on neural networks algorithms for anomaly detection \cite{kwon2017survey} and specifically on autoencoders.

\subsection{Autoencoders}
Autoencoders are deep neural networks used for the efficient encoding and reconstruction of input in an unsupervised manner.
Traditionally, autoencoders were used for dimensionality reduction and feature learning \cite{hinton2006reducing}, but currently they are also used for de-noising \cite{lu2013speech}, building generative models \cite{bengio2013generalized}, and adversarial training \cite{mescheder2017adversarial}.

\begin{figure}[h!]
  \centering
  \includegraphics[width=0.40\textwidth]{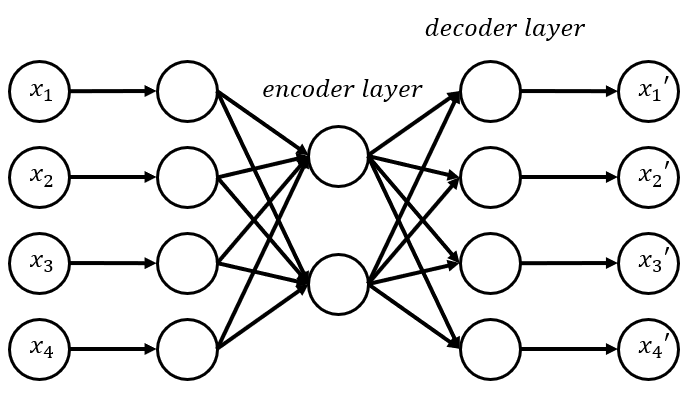}
  \caption{An example of an autoencoder with one hidden layer}
  \label{fig:autoencoder}
\end{figure}

Autoencoders consist of two components: the \textit{encoder} and the \textit{decoder} (see example in Figure \ref{fig:autoencoder}). 
The two components are trained together: the encoder compresses the data, and the decoder attempts to reconstruct it. 
The network uses the reconstruction error to adjust the weights of the network and obtain compact representations that capture the "essence" of the analyzed data.

The ability of autoencoders to reconstruct and de-noise data makes them useful anomaly detectors \cite{meidan2018n,mirsky2018kitsune,fan2018video}, particularly in cases of one-class anomaly detection (i.e. when only "normal" samples are available) \cite{erfani2016high,wei2018anomaly}. 
The autoencoder receives a sample (which can also be made "noisy" using dropout or a similar technique), compresses it using the encoder and then attempts to reconstruct the original sample using the decoder. 
The discrepancy between the original and reconstructed samples is captured by the loss function and is used to train the neural net. 
Once the network is trained, samples with high discrepancy (i.e., highly different than expected) are flagged as anomalies. 
One of the common means of measuring the discrepancy between samples is the root mean squared error (RMSE), which is calculated as 

\begin{equation}
RMSE(X, X') = \sqrt{\frac{1}{n}\Sigma_{i=1}^{n}{\Big(X_i -X'_i\Big)}}
\end{equation}

\noindent where $X$ and $X'$ are the vectors of the original and reconstructed samples, respectively.




\subsection{Generative Adversarial Nets}
Generative adversarial nets (GANs) \cite{goodfellow2014generative} are deep neural networks architectures consisting of two sub-networks: a generator and a discriminator. 
These sub-networks compete in a Nash equilibrium (zero sum game), where the goal of the discriminator is to discern samples produced by the generator from those sampled from the actual data and the goal of the generator is to fool the discriminator.
Since their introduction in 2014, GANs have been used in multiple domains, including images, music, text generation, and anomaly detection. 

The application of GANs in anomaly detection has been proposed for video \cite{ravanbakhsh2017training} and medical images \cite{schlegl2017unsupervised}. 
In the former study, the GAN was trained on the RGB channels of "normal" videos in an attempt to reconstruct corrupted videos. 
The size of the reconstruction error was used to identify anomalous section in the video. 
In the latter study, the architecture was trained on benign retina scans.
It is important to note that these studies do not require the samples generated by the GAN to be valid, as is the case in some domains (see  the Introduction section).


\section{PROPOSED METHOD}
\subsection{The proposed architecture}

Combining GANs and autoencoders requires us to reconcile two seemingly opposing goals: (1) generating samples that can be reconstructed by the autoencoder with high accuracy, and (2) generating samples that are similar to "real" data. 
The reason these two goals are contradictory is that reducing reconstruction error is easier to achieve with simplistic samples, while generating samples that are similar to the real data requires them to be more complex.

\begin{figure}[h!]
\centering
\includegraphics[scale=0.37]{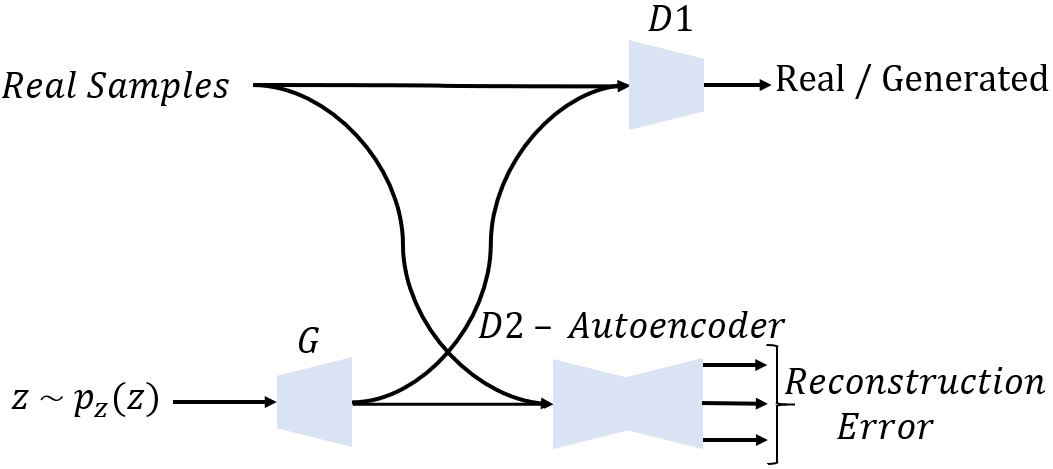}
\caption{\MethodName architecture}
\label{fig:GANarchitecture}
\end{figure}

In order to address this challenge, we propose the Multi-Discriminator Generative Adversarial Network (\MethodName). 
The proposed architecture, presented in Figure \ref{fig:GANarchitecture}, consists of a single generator $G$ and two discriminators: $D1$ and $D2$. 
While each discriminator receives, in turn, two batches of samples -- "real" and generated -- their loss functions (i.e., goals) are different:

\begin{itemize}
\item $D1$ is a feedforward network whose aim is to correctly separate the "real" samples from the generated ones. 
The goal of $G$ with respect to $D1$ is to make the two groups indistinguishable. 
We do this by defining the following two-player minimax game, represented by the following formula \cite{goodfellow2014generative}:

\begin{equation}
\label{eq1:D1}
\begin{aligned}
\minBelow_{G}\maxBelow_{D_1}V(D1, G)  = 
\mathbb{E}_{x_ \texttt{\char`\~} {p_{data^{(x)}}}} [log D_1(x)] + \\\mathbb{E}
_{z_ \texttt{\char`\~} {p_{z^{(z)}}}}[log(1- D_1(G(z)))] 
\end{aligned}
\end{equation}

where $D_1$($x$) denotes the probability assigned by $D1$ to sample $x$ of being "real" (i.e. not generated by $G$).

\item $D2$ is an autoencoder component. 
Its goal is to correctly reconstruct all the samples it receives, regardless of them being real or generated.
Unlike other GAN architectures, the goal of $G$ in this context is to assist $D2$ (i.e., reduce $D2$'s loss function values).
The loss function for \textit{both} $D2$ and $G$ is the same and is given the mean square error (MSE) of the sample reconstruction:

\begin{equation}
\label{eq1:D2}
    \ \mathcal{L} (x, x') = \mathcal{L} (D2) = \mathcal{L} (G) =   ||x-x'||^2\
    \end{equation}

\end{itemize}

\noindent We train the architecture in the following manner. 
In every iteration, $G$ generates a batch of samples. 
We send the generated batch to one of the discriminators, along with an equal size batch of real samples. 
We calculate the loss for the generator and the relevant discriminator using either formula \ref{eq1:D1} or \ref{eq1:D2}, and then use back propagation to update the networks. 
The process is then repeated with the second discriminator. 
In each iteration, the parameters of the non-participating discriminator are frozen.

\subsection{Training and initialization strategies}

\MethodName utilizes two discriminators with the goal of generating instances that are both valid and have the ability to be successfully reconstructed by an autoencoder. 
However, we considered the possibility that because of the generator only being adversarial to $D1$, the $D2$ autoencoder may be "thrown off" by samples generated in the early epochs. 
This concern stems from the fact that the generated samples of early training epochs are not likely to resemble the real data. By trying to reconstruct these wholly-unrelated samples, the autoencoder may assimilate false patterns.

To test the hypothesis that the samples generated during the initial training epochs may be detrimental to $D2$, we defined a "warm-up" period for \MethodName. 
During this period we train $G$ and $D1$ as described above, but only train $D2$ on the real samples. 
In our experiments we evaluated the performance of our model on different numbers of warm-up epochs, ranging from zero to six. 
Once the warm-up period is over, we train both discriminators on real and generated data. 
The proposed algorithm is presented in Algorithm \ref{alg:gans}.
 

\begin{algorithm}
\caption{MDGAN training}\label{alg:gans}
\begin{algorithmic}[1]
\Procedure{fit}{$real \ data, \ warm \ up \ time$}
\For{\texttt{number of training iterations}}
\State ${real \ batch} \leftarrow {sample \ batch \ from \  real \ data}$
\State optimize $D1$ on ${real \ batch}$
\State sample:  \[z \sim \mathcal{N}({0},\,{1})\,\] 
\State $generated \ data \leftarrow G(z)$
\State optimize $D1$ on $generated \ data$
\State optimize $G$ on $D1$
\State optimize $D2$ on ${real \ batch}$
\If{iteration number \ \textgreater \  ${warm \ up \ time}$}
  \State optimize $D2$ on $generated  \ data$
\EndIf
\State optimize $G$ on $D2$

\EndFor

\State \textbf{return} $D2 $
\EndProcedure
\end{algorithmic}
\end{algorithm}





\section{EVALUATION}

\begin{figure}	
	\centering
	\includegraphics[height=0.35\textheight]{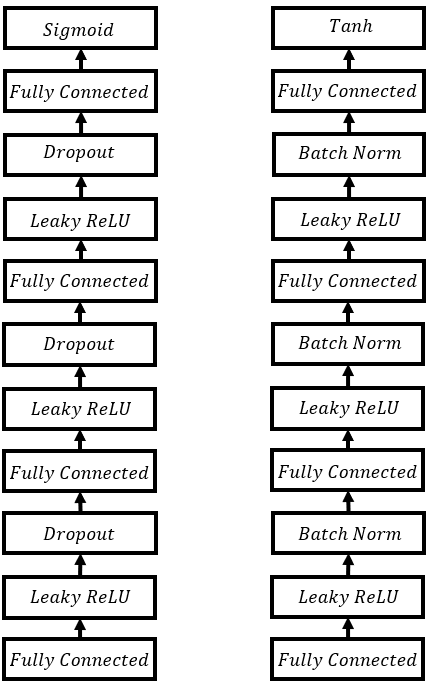}
	\caption{$D1$ architecture (left), $G$ architecture (right)}
    \label{fig:d1g}
\end{figure}

\subsection{Datasets}

We evaluated our approach on ten diverse datasets varying in size, number of attributes, and class imbalance.
The datasets are available on the OpenML\footnote{www.openML.org} and Outlier Detection DataSets (ODDS)\footnote{http://odds.cs.stonybrook.edu/} repositories and their properties are presented in Table \ref{fig:t1}.
Five datasets are well-known benchmarks for the anomaly detection task: NSL-KDD \cite{revathi2013detailed}, Pendigit \cite{keller2012hics}, Annthyroid \cite{abe2006outlier}, SWaT \cite{goh2016dataset} and breast cancer \cite{mangasarian1990cancer}.

All datasets represent binary classification problems\footnote{Some datasets were originally multi-class, but binary versions exist in online repositories}, with the minority class instances defined as the anomalies we aim to detect.

We partitioned each dataset into training, validation, and test sets.
The partitioning process varied depending on whether or not pre-defined partitions were in existence:
\begin{itemize}
\item For datasets with pre-defined partitions, we removed all anomalous samples (i.e., the minority class) from the training set.
Of the remaining training set samples, 10\% was randomly selected as the validation set. 
The test set was not changed.
\item For datasets without pre-defined partitions, we first assigned all anomalous samples to the test set. 
We then assigned the remaining samples in the manner described in Table \ref{fig:t1}. 
Again, 10\% of the training set was randomly assigned to the validation set.

\end{itemize}

\begin{table}[h]
\centering
\small
\begin{tabular}{|p{1.60cm}|p{1.2cm}|p{1.23cm}|p{1.23cm}|p{1.25cm}|}
	\hline
	\textbf{Dataset} & \textbf{\# of Features} & \textbf{Trainset size}  & \textbf{Testset size} & \textbf{Anomalies (\%)} \\
	\hline
	NSL-KDD* & 39 & 67,343 & 22,544 & 56.92 \\
	\hline
	Pendigit \cite{keller2012hics} & 16 & 6,000 & 1,870 & 17.90 \\
	\hline
	Video injection \cite{mirsky2018kitsune} & 115 & 1,000,000 & 1,369,902 & 6.96 \\
	\hline
    Annthyroid \cite{abe2006outlier} & 6 & 6,000 & 1,200 & 44.50 \\
	\hline
	Forest cover \cite{liu2008isolation} & 10 & 250,000 & 36,048 & 7.62 \\
	\hline
	Breast cancer \cite{mangasarian1990cancer} & 10 & 200 & 599 & 48.29 \\
	\hline
	CPU & 18 & 3,000 & 5,192 & 47.70 \\
	\hline
    Ailerons & 40 & 2,000 & 11,750 & 49.60 \\
	\hline
    SWaT* \cite{goh2016dataset} & 51 & 496,000 & 449,919 & 11.98 \\
	\hline
	Yeast \cite{Dua:2017} & 8 & 1,014 & 470 & 7.74 \\
    \hline

\end{tabular}
\caption{The characteristics of the evaluated datasets (number of features, size of the training and test sets, and the percentage of anomalies). "*" indicates datasets with pre-existing partitions}
\label{fig:t1}
\end{table}

\noindent In addition, we normalized all numeric features to the range [-1,1] and removed all categorical features with more than three values from the datasets.
We took the latter action, because the categorical features had to be represented using sparse vectors, and this resulted in reduced performance for both \MethodName and the baseline.

\begin{table}
\centering
\small
\begin{tabular}{|p{1.9cm}|p{1.15cm}|p{1.15cm}|p{1.15cm}|p{1.15cm}|}
	\hline
	\textbf{Dataset} & \textbf{No \ Warm \ Up}  & \textbf{One Epoch\ Warm\ Up} & 	
    \textbf{Three Epochs\ Warm\ Up} &  \textbf{Six Epochs\ Warm\ Up} \\
	\hline
	NSL-KDD & 0.43\% & 0.9\%* & 0.51\% & 0.66\%* \\
	\hline
	Pendigit &  4.25\%* & 2.77\% &  3.42\%* & 1.48\% \\
	\hline
	Video injection & 0.03\% & -2\%* & -0.4\% & -0.63\%*  \\
	\hline
    Annthyroid & -3.96\%* & -4.01\%* & -5.9\%*  & -5.03\%*  \\
	\hline
	Forest cover & 6.73\%* & 4.12\% & 1.29\% & 1.77\% \\
	\hline
	Breast cancer & 5.53\%* & 5.52\%* & 4.92\%* & 3.78\%* \\
	\hline
	CPU & 0.78\%* & 0.88\%* & 0.66\%* & 0.69\%* \\
	\hline
    Ailerons & 2.83\%* & 1.78\%* & 2.21\%* & 1.76\%* \\
	\hline
    SWaT &  0.57\%  & 2.27\%  & 2.65\%  & 2.49\%  \\
	\hline
	Yeast & -2.86\%  & -5.57\%*  & -2.77\%  & -3.44\%  \\
	\hline
\end{tabular}
\caption{Percentage of improvement in AUC against the baseline (higher is better), averaged by 30 different seeds.
"*" indicates significance with 95\% confidence}
\label{fig:tauc}
\end{table}

\subsection{Training parameters}
We used the following settings throughout the evaluation:
\begin{itemize}
\item \textbf{Stopping criteria} -- all models were trained for 30 epochs. 
We then chose the architecture configuration that was in place for the epoch with the highest score on validation set.

\item \textbf{Learning rate and optimizers} -- $D1$ was optimized using a stochastic gradient descent optimizer with a learning rate of 0.01. 
$D2$ and $G$ were optimized using the Adam optimizer with a learning rate of 0.001.

\item \textbf{Dropout and batch normalization} -- $G$ contains a 10\% rate dropout after each hidden layer. 
$D1$ contains batch normalization after each hidden layer.

\item \textbf{Warm up values} -- we evaluated $D2$ with warm up values of zero, one, three, and six epochs (see "Training and initialization strategies" in the previous section for more details).

\item \textbf{Initialization} -- each experiment was run 30 times, using different initialization parameters.

\end{itemize}

\subsection{Experimental setting}

\noindent \textbf{The baseline.} Since the goal of \MethodName is to improve the performance of an autoencoder through the generation of additional samples, we compared our approach to an autoencoder with an identical configuration to the one used by our $D2$ component. The same validation set-based stopping criteria was also applied. \newline

\begin{table}[t]
\centering
\small
\begin{tabular}{|p{1.9cm}|p{1.15cm}|p{1.15cm}|p{1.15cm}|p{1.15cm}|}
	\hline
	\textbf{Dataset} & \textbf{No \ Warm \ Up}  & \textbf{One Epoch\ Warm\ Up} & \textbf{Three Epochs\ Warm\ Up} &  \textbf{Six Epochs\ Warm\ Up} \\
	\hline
	NSL-KDD & 0 & 0.3\% & 0.3\% & 0.3\%* \\
	\hline
	Pendigit & 14.2\%* & 12.6\%* &  12.5\%* & 3.9\% \\
	\hline
	Video injection & -0.01\%* & -0.01\%* & 0 & -0.22\% \\
	\hline
    Annthyroid & -2.4\% & 2.3\% & 4.1\%* & 4.2\%* \\
	\hline
	Forest cover & 44.8\%* & 25.2\%* & 25.6\% & 12.3\% \\
	\hline
	Breast cancer & 3.3\%* & 3.3\%* & 3\%* & 2.3\%* \\
	\hline
	CPU & 0.7\%* & 0.8\%* & 0.6\%* & 0.6\%* \\
	\hline
    Ailerons & 3.3\%* & 2.4\%* & 2.6\%* & 2.1\%* \\
	\hline
    SWaT & -0.04\% & 7.6\% & 2.2\% & 4.7\% \\
	\hline
	Yeast & -5.1\% & 6.6\% & 4.3\% & 5.8\% \\
	\hline

\end{tabular}
\caption{Percentage of improvement in AUC PR against the baseline (higher is better), averaged by 30 different seeds. "*" indicates significance with 95\% confidence}
\label{fig:taucpr}
\end{table}

\begin{table}[t]
\centering
\small
\begin{tabular}{|p{1.9cm}|p{1.15cm}|p{1.15cm}|p{1.15cm}|p{1.15cm}|}
	\hline
	\textbf{Dataset} & \textbf{No \ Warm \ Up}  & \textbf{One Epoch\ Warm\ Up} & \textbf{Three Epochs\ Warm\ Up} &  \textbf{Six Epochs\ Warm\ Up} \\
	\hline
	NSL-KDD & 0 & -2.5\% & -3.9\%* & -5.3\%* \\
	\hline
	Pendigit &  -20.6\%* & -12.8\% &  15.7\%* & 8.6\% \\
	\hline
	Video injection & 1.1\% & 7.7\%* & 4.1\%* & 3.3\%* \\
	\hline
    	Annthyroid & 6.2\%* & 5.7\% & 8.1\%* & 7.5\%* \\
	\hline
	Forest cover & -15.9\%* & -10.8\%* & -5.8\% & -4.3\% \\
	\hline
	Breast cancer & -10.8\%* & -12.4\%* & -10.2\%* & -8.5\%* \\
	\hline
	CPU & -5.01\%* & -5.5\%* & -4.4\%* & -4.2\%* \\
	\hline
    	Ailerons & -6.3\%* & -4.2\%* & -4.8\%* & -3.7\%* \\
	\hline
    	SWaT & 1.5\% & -0.1\% & -2.9\% & -1.7\% \\
	\hline
	Yeast & 4.9\% & 11.4\% & 6.5\% & 9\% \\
	\hline

\end{tabular}
\caption{Percentage of improvement in EER against the baseline (lower is better), averaged by 30 different seeds. "*" indicates significance with 95\% confidence}
\label{fig:teer}
\end{table}

\noindent \textbf{Evaluation measures.} We used three evaluation measures to analyze our results:
\begin{itemize}
\item \textbf{Area under the receiver operating characteristic curve (AUC)} -- used to measure the performance of our approach across all possible true-positive/false-positive values.

\item \textbf{Area under the precision-recall curve (AUC PRC)} -- considered to be more informative than AUC when the percentage of anomalies is low (i.e., imbalanced datasets). 

\item \textbf{Equal error rate (EER)} -- designed to measure the trade-off between the false-positive and false-negative rates.
\end{itemize}

\noindent \textbf{Statistical tests.} To validate the significance of our results, we used the paired t-test on the three evaluation measures described above. 
We marked results as significant in cases where the confidence level of 95\% or higher. \newline

\noindent The architectures used in our experiments are presented in Figure \ref{fig:d1g}.
Their structure is as follows: $G$ is a four layer neural net. We used the leaky ReLU function with alpha 0.2 after each hidden layer and batch normalization. 
Finally, after the last fully connected layer we apply the tanh activation function.

$D1$ is also a four layer neural net. 
We used the leaky ReLU function with alpha 0.2 after each hidden layer, and then we applied dropout. 
Finally, after the last fully connected layer we applied the sigmoid activation function to classify as real or fake.

$D2$ is a four layer autoencoder, encoding to 70\% of the input dimension, and then to 50\%. The decoder is the exact opposite, decoding to 70\% and then to a 100\%, same as the input dimension. We apply ReLU after each hidden layer, and tanh after the output layer.

\begin{figure}[t]
  
  \centering \hspace*{-0.5cm}
  \includegraphics[height=0.9\textheight, width=0.502\textwidth , keepaspectratio]{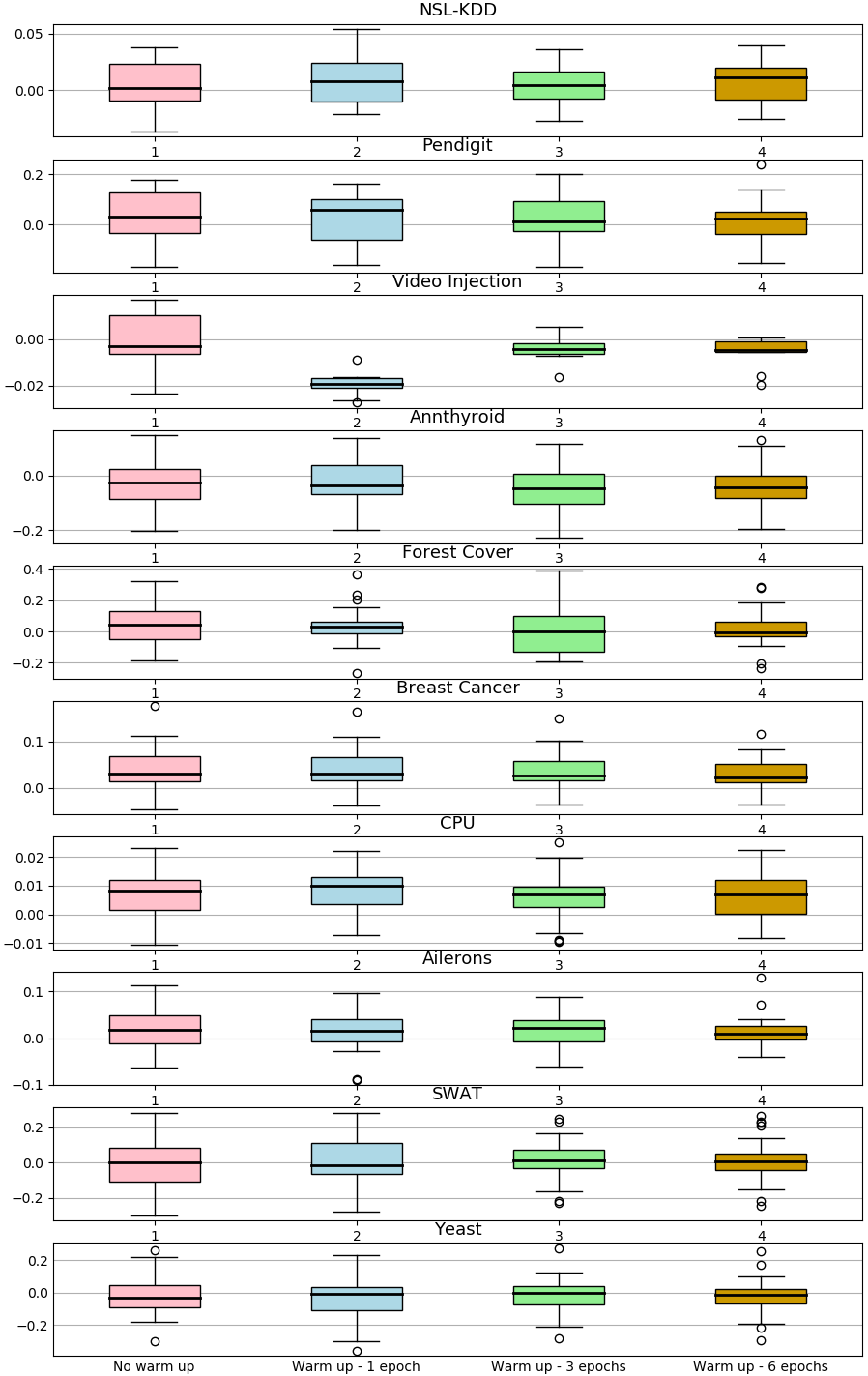}
  \caption{Box-and-whisker plot of the differences in AUC between the various warm-up configurations and the baseline, across 30 different seeds}
  \label{fig:whisker}
\end{figure}

\subsection{Results}

\begin{figure*}[t]
  \label{fig:AUCimprov}
  \centering
  \includegraphics[scale=0.51]{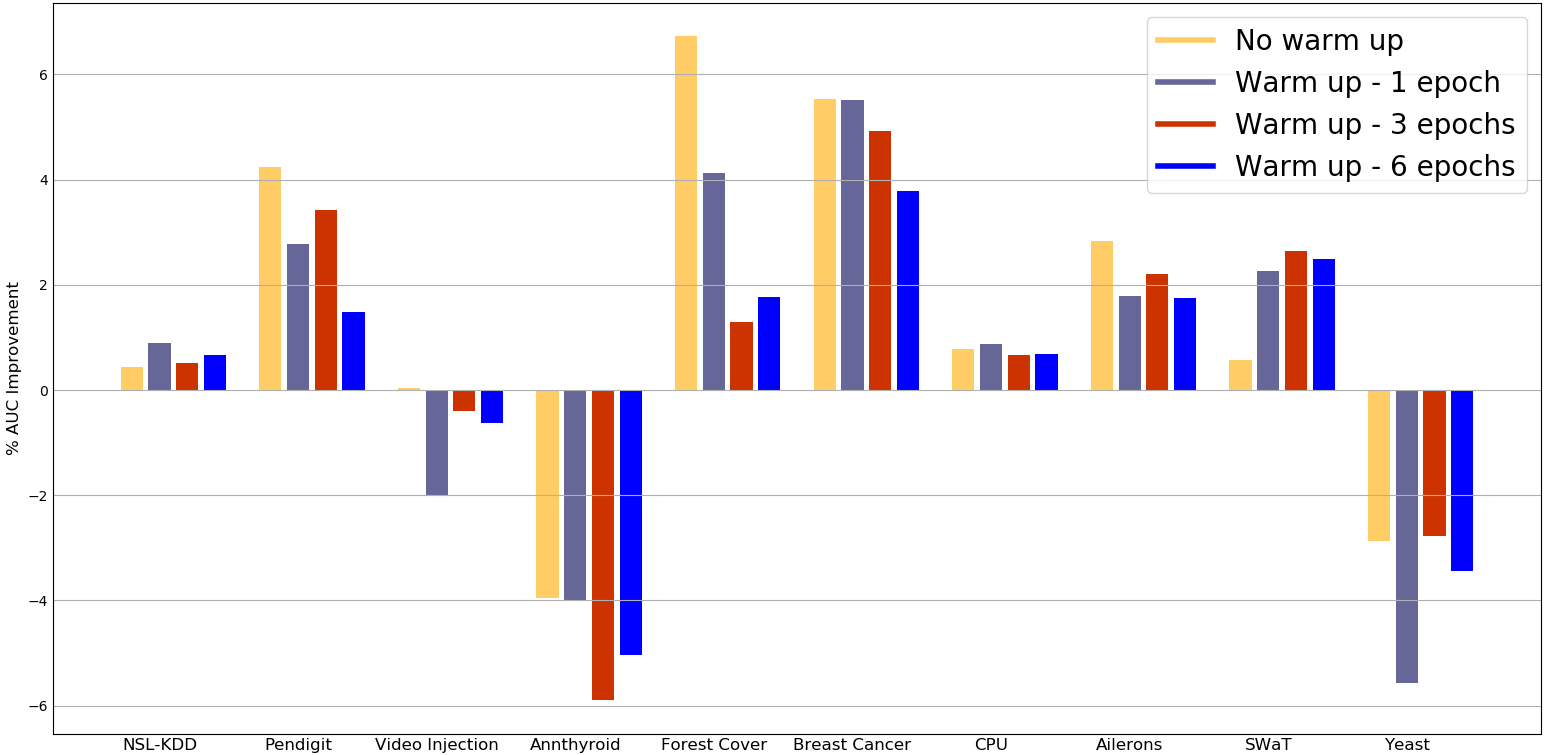}
  \caption{\% AUC improvement of the different warm ups to the baseline, averaged by 30 different seeds}
  \label{fig:barplot}
\end{figure*}

Tables \ref{fig:tauc}--\ref{fig:teer} and Figures \ref{fig:whisker} and \ref{fig:barplot} present the performance of \MethodName according to the three computed performance measures. 
In each table we present the ratio of the performance measure (averaged over the 30 experiments) of the \MethodName autoencoder $D2$ and the baseline autoencoder. 

From the results it can be observed that seven of the ten datasets, the GAN autoencoder outperformed the baseline autoencoder with the difference being statistically significant except for one dataset ($SWaT$).

Another interesting (and perhaps counter-intuitive) observation is that the warm up period does not always improves the performance of the GAN autoencoder (and even sometimes leads to a degradation in the performance).
From our analysis of the data we conclude that the samples generated during the warm up period are too similar to the real samples, thus reducing the generator's ability to generate "unexpected" (and valuable) samples that would make the autoencoder more robust. 
See figure \ref{fig:AUCimprov} for a comparison of the warm-up strategies.

Finally, our results lead us to conclude that the size of the training set does affect the performance of our approach (relative to the baseline). However, the \textit{complexity of the problem }(i.e., the number of features per sample) has a clear effect on the results: the two datasets for which \MethodName fails to improve are those with the smallest number of features. This leads us to conclude that our approach is better suited for a high-dimensional space, where the generation of additional samples is a more challenging task.

\begin{figure}[h]
  \centering
  \includegraphics[scale=0.26]{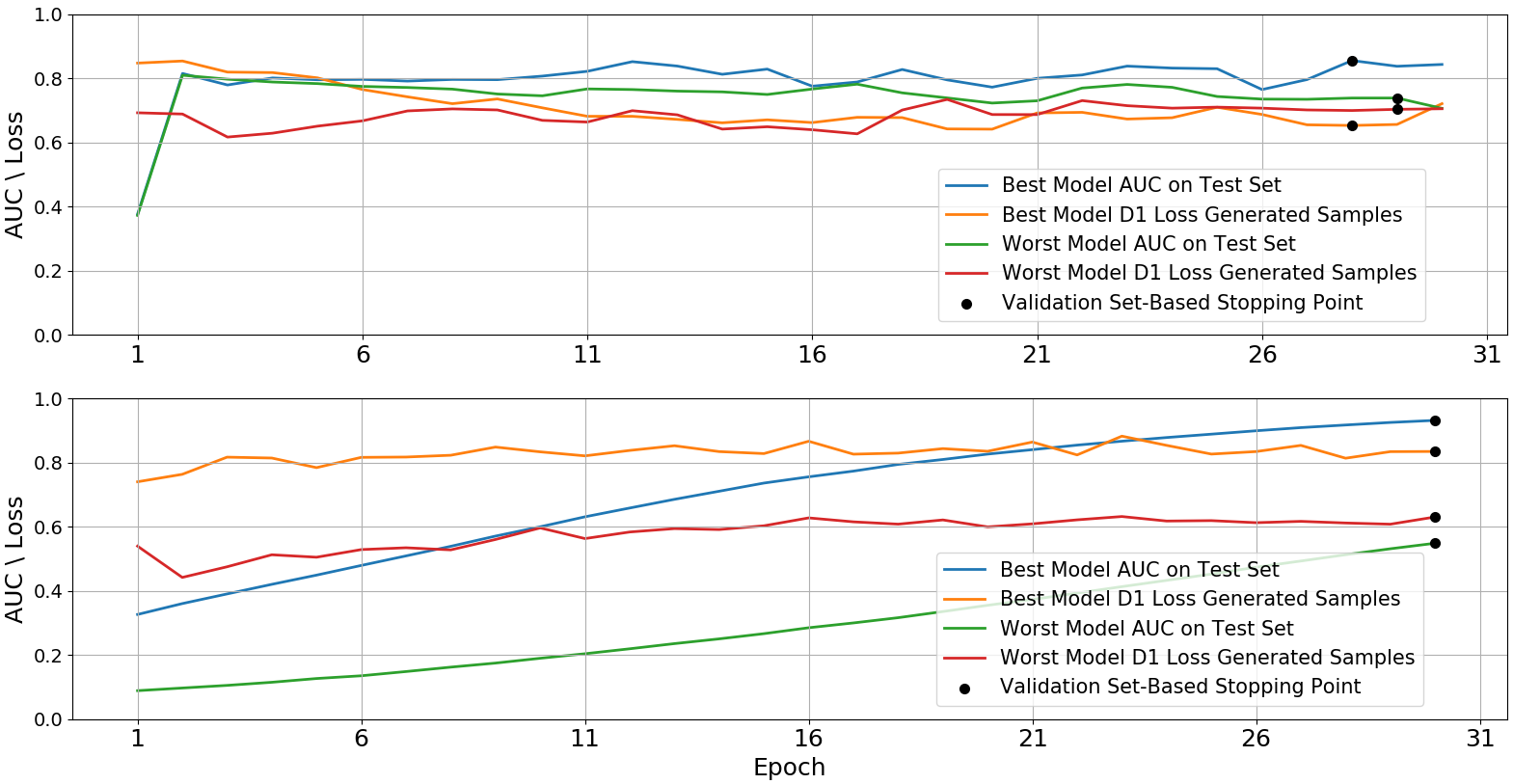}
  \caption{$D1$ loss on generated samples during training, best and worst performance runs; Ailerons (top), Breast Cancer (bottom). It can be seen that the top-performing architectures have higher $D1$ loss values.}
  \label{fig:ailerons_breast}
\end{figure}

\begin{figure}[h]
  \centering
  \includegraphics[scale=0.26]{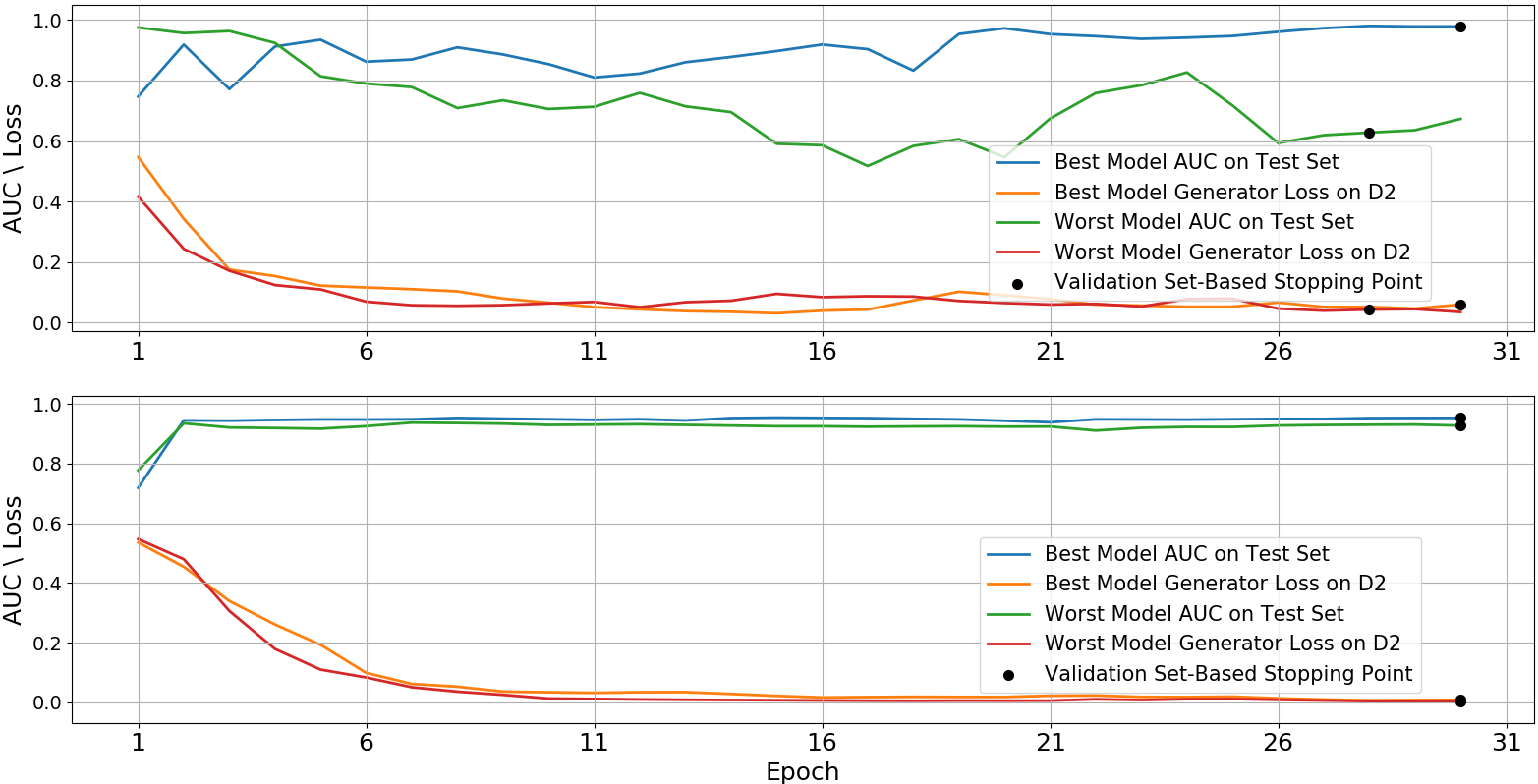}
  \caption{$G$ loss on $D2$ during train, best and worst performance runs; Pendigits (top), CPU  (bottom). It is clear that the loss converges more slowly for the top-performing architectures.}
  \label{fig:pendigits_cpu}
\end{figure}

\section{DISCUSSION}
We analyzed the performance of the various components (i.e., the generator and the two discriminators) and were able to draw the following conclusions. \newline

\noindent \textbf{Early "peaking" is indicative of lower performance.} For each of our datasets, we compared the two runs with the best and worst results for the ``no warm-up'' configuration. In six out of the ten datasets, the run which achieved the lower AUC score terminated considerably earlier (3.5 epochs earlier, on average) when using the validation performance as a stopping criteria. In the four other datasets, both the best and the worst were trained for the full 30 epochs (the maximal number).

We hypothesize that these results reflect the fact that our \MethodName architecture requires a longer training time to perform well due to the fact that it has to satisfy a larger number of constraints compared with a "standard" GAN architecture. 
A short training time may be indicative of the GAN producing samples do not contribute to the training process. \newline
\noindent  

\textbf{The dense discriminator functions as a regulator.} One of the base hypothesis behind \MethodName is that the dense discriminator network ($D1$) will assist in guiding the generator towards generating more "valid" samples.
In order to test this hypothesis, we compared the loss function of $D1$ on the \textit{generated samples only} for the best and worst-performing runs in each dataset. 
Our results show direct correlation between higher loss function values for $D1$ and improved overall performance. 
An example of this is presented in Figure \ref{fig:ailerons_breast}. 
\newline 



\textbf{Slower generator-autoencoder convergence is indicative of better results.} We once again compare the best and worst performance for each dataset, but this time we focus on the \textit{generator loss} with respect to each discriminator. 
Our analysis shows that while in most cases the loss for $D1$ (the dense DNN) is similar for the best and worst cases, the top-performing scenarios often displayed higher initial generator-loss when training on the autoencoder. 
Moreover, the loss reduction was noticeably slower. 
An example of this scenario is presented in Figure \ref{fig:pendigits_cpu}. 


We argue that these results show that are model deploys a variant of adversarial training, in the sense that \MethodName performs better when the autoencoder has greater difficulty reconstructing the generated samples. 
We find it likely that as a result, the autoencoder becomes better at reconstructing previously-unseen types of samples.

\section{CONCLUSIONS AND FUTURE WORK}

In this study we have presented \MethodName, a novel multi-discriminator GAN approach for anomaly detection. 
Our architecture enables the GAN to reconcile two conflicting aims: 1) generating sophisticated samples that can pass "real" instances of the dataset, and; 2) create instances that can be accurately reconstructed by an autoencoder. Using our approach, we have been able to improve the performance across a variety of datasets.

For future work, we plan to pursue several directions. First, we will experiment with more advanced training strategies, dynamically allocating different weights to each discriminator. Secondly, we will explore architectures with a larger number of discriminators in an attempt to reconcile a more complex of constraints. Thirdly, we will evaluate our method in the context of generating adversarial samples. Finally, we plan to apply our approach to additional domains.

\bibliographystyle{aaai}
\bibliography{ref}

\end{document}